# MOSI: Multimodal Corpus of Sentiment Intensity and Subjectivity Analysis in Online Opinion Videos

**Amir Zadeh**
Carnegie Mellon University, Pittsburgh, PA 15213
abagherz@cs.cmu.edu

**Rown Zellers**
University of Washington, Seattle, 98195
rzellers@hmc.edu

**Eli Pincus**
University of Southern California, Los Angeles, CA 90089
pincus@ict.usc.edu

**Louis-Philippe Morency**
Carnegie Mellon University, Pittsburgh, PA 15213
morency@cs.cmu.edu

## Abstract

People are sharing their opinions, stories and reviews through online video sharing websites every day. Studying sentiment and subjectivity in these opinion videos is experiencing a growing attention from academia and industry. While sentiment analysis has been successful for text, it is an understudied research question for videos and multimedia content. The biggest setbacks for studies in this direction are lack of a proper dataset, methodology, baselines and statistical analysis of how information from different modality sources relate to each other. This paper introduces to the scientific community the first opinion-level annotated corpus of sentiment and subjectivity analysis in online videos called Multimodal Opinion-level Sentiment Intensity dataset (MOSI). The dataset is rigorously annotated with labels for subjectivity, sentiment intensity, per-frame and per-opinion annotated visual features, and per-milliseconds annotated audio features. Furthermore, we present baselines for future studies in this direction as well as a new multimodal fusion approach that jointly models spoken words and visual gestures.

## 1 Introduction

Video sharing websites such as YouTube, Vine, and Vimeo are increasingly popular. For example, more than 300 hours of video are uploaded every minute to YouTube[1]. Many people share their opinions, stories and reviews through these online video postings. This phenomenon has seen a growing attention from many companies, researchers and consumers interested in building better opinion-mining applications for summarization, question answering and video retrieval. We highlight bellow three main challenges of studying sentiment in these online opinion videos.

First challenge comes from the volatile and high-tempo nature of these opinion videos where speakers will often switch between topics and opinions. This makes it challenging to identify and segment the different opinions expressed by the speakers. For example a speaker can express more than one opinion in the same spoken utterance, as in: "*that was a great effect, there is a lot of cheap childish humor everyone can relate to but I thought it was hilarious*".

The second challenge comes with the range and subtlety of sentiment intensities expressed in these opinion videos. We want approaches not only able to recognize the polarity of a video segment (e.g., positive or negative) but also estimate the strength of the expressed sentiment.

Third challenge is a fundamental research question on how to use information more than text. To simply focus on the spoken words (e.g., text-based sentiment analysis) may bring an ambiguity that would be resolved with the visual information. For example, the sentiment expressed by the spoken opinion "*this movie was different*" may not be clear by itself but observing a strong frown expression from the speaker can make it clear that it is negative.

---
[1] http://www.youtube.com/yt/press/statistics.html

In this paper we introduce a novel corpus for studying sentiment and subjectivity in opinion videos from online sharing websites such as YouTube[2]. To address the first challenge, we present a subjectivity annotation scheme for fine-grained opinion segmentation in online multimedia content. 3702 video segments were reliably identified in our MOSI dataset using this scheme, including 2199 opinion segments. Sentiment in each opinion segment was annotated as a spectrum between highly positive and highly negative to address the second challenge. As enabling steps toward the third challenge, we present a multimodal study of language and gesture related to sentiment intensity that leads to the idea of multimodal dictionary. We also make available, as part of the MOSI dataset, transcriptions that were carefully synchronized with acoustic and visual features at both word and phoneme level, to ensure the usability of dataset for future multimodal studies of language.

The following section presents related work on text-based sentiment analysis and multimodal analysis. Section 3 introduces our new dataset for opinion-level multimodal sentiment analysis as well as primary analysis of multimodal data to find interaction patterns between words and gestures. Section 4 presents baseline for sentiment and subjectivity studies. We conclude this paper with Section 5.

## 2 Related Work

This work is connected to areas in natural language processing, multimodal analysis and emotion recognition.

**Multimodal sentiment analysis datasets** YouTube Opinion Dataset created by Morency et.al. (2011) is a dataset for multimodal analysis of sentiment. It contains 47 videos from YouTube annotated for sentiment polarity at video level by three workers. The dataset consists of manually transcribed text and automatically extracted audio and visual features, as well as automatically extracted utterances. MMMO dataset (Wollmer et.al. 2013) is an extension of YouTube Opinion Dataset that extends the number of videos from 47 to 370. Spanish Multimodal Opinion Dataset created by Rosas et.al. (2013) is a Spanish multimodal sentiment analysis dataset. It consists of 105 videos annotated for sentiment polarity at utterance level. Utterances are extracted automatically based on long pauses with most videos having 6-8 utterances. The dataset contains 550 utterances in total. None of the proposed datasets have sentiment intensity annotations; they rather focus on polarity. Also they mostly focus on analysis of videos or utterances rather than fine grained analysis of sentiment as mentioned in introduction.

**Text based sentiment analysis** Research in text based sentiment analysis has been an active and extremely successful field (Pang and Lee, 2004, Vinodhini et.al., 2012). Among notable efforts are works done in automatically identifying opinion words and their sentiment polarity (Hatzivassiloglou and McKeown, 1997; Turney, 2002; Hu and Liu, 2004; Taboada et al., 2011), studies using n-grams and more complex language models (Takamura et al., 2006; Yang and Cardie, 2012) and works addressing sentiment compositionality by using polarity shifting rules or careful feature engineering (Polanyi and Zaenen, 2006; Moilanen and Pulman, 2007; Rentoumi et al., 2010; Nakagawa et al., 2010). These methods have been used in many different applications including opinion mining in tweets and online forums (Pak et.al. 2010; Argawal et.al. 2011; Balog et al., 2006), analysis of political debates (Carvalho et al., 2011), question answering (Oh et al., 2012), conversation summarization (Carenini et al., 2008), and citation sentiment detection (Athar and Teufel, 2012). Recursive Neural Networks have also been successful in sentiment analysis. Socher et.al (2013) addressed sentiment analysis as a 5 class classification task and managed to get accuracy of 45.7% using Recursive Neural Tensor Networks. Their results show that sentiment intensity analysis is far from solved for text. All these approaches are primarily focusing on the (spoken or written) text, ignoring the other communicative modalities which are helpful when analyzing videos.

**Multimodal analysis** Some earlier work introduced acoustic and paralinguistic features to the text-based analysis for the purpose of subjectivity or sentiment analysis (Somasundaran et al., 2006; Raaijmakers et al., 2008; Mairesse et al., 2012; Metze et al., 2009). Closer to our current study is the work reported in (Morency et al., 2011; Perez-Rosas et al., 2013), where multimodal cues, including visual cues, have been used for the sentiment analysis in product and movie reviews. Their approach directly concatenated modalities in an early fusion representation, without studying the relations between different modalities. Their experiments were also performed in a speaker-dependent manner, with no analysis of sentiment intensity. Poria et.al. (2015) used convolutional neural networks for multimodal sentiment analysis. However, their experiments also focused on utterances rather than opinion segments and their approach was speaker dependent with focus on sentiment polarity rather than intensity.

---
[2] Videos under creative commons license can be edited and used without the need for authors consent.

**Audio-visual emotion recognition** Among the related works are research done in the fields of affective computing and computer vision to detect emotions based on visual cues. Facial expressions are among the most powerful means of communicating emotions and intentions in human interactions (Tian et al., 2001). Interesting studies discuss findings about mental state of the speaker based on facial expressions, head gestures and other visual cues (El Kaliouby and Robinson, 2005; Baltrusaitis et.al., 2011; Calder et al., 2001; Rosenblum et al., 1996). More recently, many challenges have been organized focusing on the recognition of emotions using audio and visual cues (Dhall et al., 2014; Valstar et al., 2014), which brings the participation of many teams from around the world.

## 3 MOSI: Multimodal Opinion-level Sentiment Intensity Corpus

In this section, we introduce the Multimodal Opinion-level Sentiment Intensity (MOSI) dataset which contains: (1) multimodal observations including transcribed speech and visual gestures as well as automatic audio and visual features, (2) opinion-level subjectivity segmentation, (3) sentiment intensity annotations with high coder agreement, and (4) alignment between words, visual and acoustic features. The following sub-sections are describing the dataset in more details.

### 3.1 Acquisition Methodology

Videos were collected from the YouTube website with a focus on video-blogs, or vlogs: popular update videos used by many YouTube users to express opinions about many different subjects, often indexed by #vlog. One big advantage of this type of videos is that they usually contain only one speaker, looking primarily at the camera. The videos are recorded in diverse setups, some users have high-tech microphones and cameras, while others are using less professional recording devices. Users are in different distances from the camera. Background and lighting condition is variable between videos. The videos were kept in their original resolution and recorded in MP4 format. The length of the videos vary from 2-5 minutes.

A total of 93 videos were randomly selected using these guidelines. The final set of videos contained 89 distinct speakers, including 41 female and 48 male speakers. Most of the speakers were approximately between the ages of 20 and 30. Although the speakers were from different ethnic backgrounds (e.g., Caucasian, African-American, Hispanic, Asian), all speakers expressed themselves in English and the videos originated from either United States of America or United Kingdom. Sample snapshots of our MOSI dataset[3] are shown in Figure 1.

All video clips are manually transcribed to extract spoken words as well as the start time of each spoken utterance. Our transcription methodology consisted of four stages. First, an expert transcriber manually transcribed all the videos followed by a second transcriber reviewing and correcting all the transcriptions. Our transcription scheme contained details about pause fillers (umm, uhh, etc.), stresses and speech pause. In the third stage, the text was aligned at word and phoneme levels with the audio using a forced aligner called P2FA[4]. During the final stage, the results of the alignment were manually checked and if necessary corrected using PRAAT (Paul and Weenink, 2001).

### 3.2 Opinion-level Subjectivity Segmentation

As previously discussed, an important requirement of our dataset is to perform subjectivity segmentation at the opinion level so that we can achieve fine-grained sentiment analysis. Following the work of Wiebe et.al. (2005), subjective sentences are defined as expressions of one's opinions while objective sentences express facts and truth. This section presents the formal definition of our segmentation scheme which expands their work to extract spoken opinion segments[5].

---

[3] Please contact abagherz@cs.cmu.edu or morency@cs.cmu.edu for accessing the dataset.
[4] https://github.com/srubin/p2fa-vislab

[5] In the remaining of the paper, the terms subjective segment, opinion segment and opinion are interchangeably used and refer to the same concept.

We define subjectivity as an expression of a private state, one that is distinguishable by carrying an opinion, belief, thought, feeling, emotion, goal, evaluation or judgment. The following three rules are used to define subjectivity:

- Explicit mention of a private state – a direct mention of private state. Example: "*I also love the casting of Mark Strong as Sinestro.*"
- Speech events expressing private states – private state has been said or written by another. Example: "*Shia LaBeouf said that the second movie lacked um heart.*"
- Expressive subjective – not a direct opinion but implicit reference to an opinion. Example: "*I would never recommend watching this movie.*"

To more accurately annotate the boundaries of each subjective segment, following rules have been defined, where utterances are processed sequentially. If the utterance contains expression of private state (i.e. subjective content defined as three bullet items above) the following rules of segmentation apply (brackets are used to hold segments from here on):

- Segment the subjective content based on number of private states revealed. For example: "[*I love Shawshank Redemption*][*and I love transformers*]. While the whole utterance is subjective, proposed annotation scheme signals a clear distinction between the marked segments as the speaker shows an opinion about two different subjects, in this case movies. However if there are syntactic dependencies between the prospective segments, this rule will not apply. As an example consider "[*I love books and movies*]" versus "[*I love books*][*and I love movies*]". In the first sentence, it is not possible to identify the speaker's sentiment towards movies if "[*and movies*]" is considered in isolation; so here the second private state is considered to have syntactic dependence on the first private state. However, this is not the case in the second sentence and therefore the sentence is decomposed into 2 subjective segments; one for each private state.

- Segment if the utterance contains a modification of a private state while maintaining the subject. Example: "[*Well, based on what I saw today, I feel like the movie industry is going crazy*][*or maybe it's just me being so hard on the poor actors.*]". Here the speaker adds a corollary or addendum to his general feelings on the movie industry; that perhaps the feeling stems from him being "*too hard on the poor actors.*" This type of corollary or addendum is considered a modification. Another case comes up when sentence contains reasoning. This rule applies to any type of modification. Reasoning being one example, based on rules of subjectivity a new segment should be created for reason if the sentiment is also revealed in that reason; otherwise we combine the reason with the rest of the utterance into one segment. To better elaborate consider "[*I love movies because of the acting*]" vs "[*I love movies*][*because people pretending to be someone else is funny.*]". The former actually has only one opinion in it, but the latter has two; one for the person liking movies and the other one for people's acting being funny.

- Segment if subjective utterance ended by the start of an objective segment. For example: "[*In my opinion the movie was all about eating healthy food*], *you could see banners of different organic brands in several shots.*"

- If there is subjective content and the subjective content extends beyond the boundary of the utterance while retaining the opinion, we merge the extension with the original utterance (the extension can be multiple sentences or part of a sentence). For example: "*[I dont like it! It's not a likable movie!]*"

The subjectivity annotation was done by two trained annotators. The inter-annotator agreement is calculated via a discretizing measure. For each annotator we labeled each word in the transcript as a 1 if it is in a subjective segment and 0 if it is in an objective segment. This way each annotator is represented as a string of 0's and 1's. This information was then used to calculate Krippendorf's alpha (Krippendorf, 1970) to produce a final measurement of inter-coder agreement that accounts for chance agreement. The two annotations resulted in a Krippendorf's alpha of 0.68.

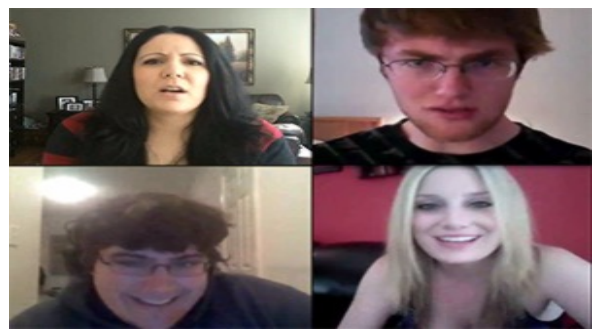

**Figure 1**: Example snapshots of videos from our new MOSI dataset.

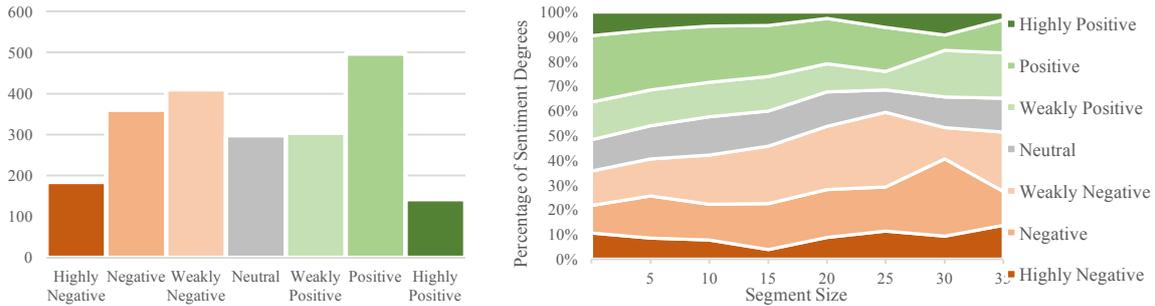

**Figure 2**: Histogram in the left shows the distribution of sentiment over the entire dataset. The right graph shows the percentage of each sentiment degree per segment size (number of words in opinion segment).

To merge the annotation results of the two annotators, an intersection approach of the two annotation sets is taken. The annotations are aligned and a segment is reported in the final set only if its content are reported as subjective by both annotators. In case the annotators have reported segments that overlap, the partial segment, from the intersection of two, is then manually extended to fit the boundaries of the sentence it occurred in. In cases that subjective segment reported by one author covers multiple segments of the other, the smaller ones are reported in the final set, because they are syntactically independent and have different opinions.

The subjectivity annotation resulted in 2199 subjective segments and 1503 objective ones. Both subjective and objective segments are considered for subjectivity studies while for sentiment annotations we only focus on subjective segments. Detailed statistics of the dataset and opinion segments can be found in Table 1.

### 3.3 Sentiment Intensity Annotation

Sentiment intensity is defined from strongly negative to strongly positive with a linear scale from -3 to +3. The intensity annotations were performed by online workers from Amazon Mechanical Turk website. Only master workers with approval rate of higher than 95% were selected to participate. A total of 2199 short video clips were created from the subjective opinion segments (see Section 3.2). For each video the annotators had 8 choices: strongly positive (labeled as +3), positive (+2), weakly positive (+1), neutral (0), weakly negative (-1), negative (-2), strongly negative (-3) and also they were given a choice "uncertain" if they were not sure.

The instructions were kept simple on purpose to reduce any bias. The only tutorial was on how to use the online system (e.g., how to submit the form). The task was phrased as following: "*How would you rate the sentiment expressed in this video segment? (Please note that you may or may not agree with what the speaker says. It is imperative that you only rate the sentiment state of the speaker, not yourself)*". Each video clip was annotated by 5 different workers. The workers were offered to annotate as many video clips as they wanted (but could not annotate the same clip twice). The inter annotator agreement between workers was 0.77 in terms of Krippendorf's Alpha. The final sentiment intensity of each segment is the average of all 5 workers. Figure 2 shows the distribution of sentiment intensities for all opinion segment in MOSI dataset on the left side. On the right side of Figure 2, we show how the sentiment distribution changes as the size of the opinion (number of words in that opinion) increases. It is interesting to see that the proportions are mostly constant across all segment sizes signaling a difference between sentiment analysis in videos and text where most of the short opinions are neutral (Socher et.al. 2013).

### 3.4 Visual Gesture Annotation

Audio and visual features have been automatically extracted from MPEG files with framerates of 1000 for audio and 30 for video. Visual features include 16 Facial Action Units, 68 Facial Landmarks, Head Pose and Orientation, 6 Basic Emotions[6] and Eye Gaze (Wood et.al., 2015, Baltrusaitis et.al., 2012, Baltrusaitis et.al., 2014). More than 32 audio features including pitch, energy, NAQ (Normalized Amplitude Quotient), MFCCs (Mel-frequency Cepstral Coefficients), Peak Slope, Energy Slope have also been extracted using COVAREP (Degottext et.al., 2014). All the audio and visual features are publicly available with the dataset.

---
[6] http://imotions.com/products/

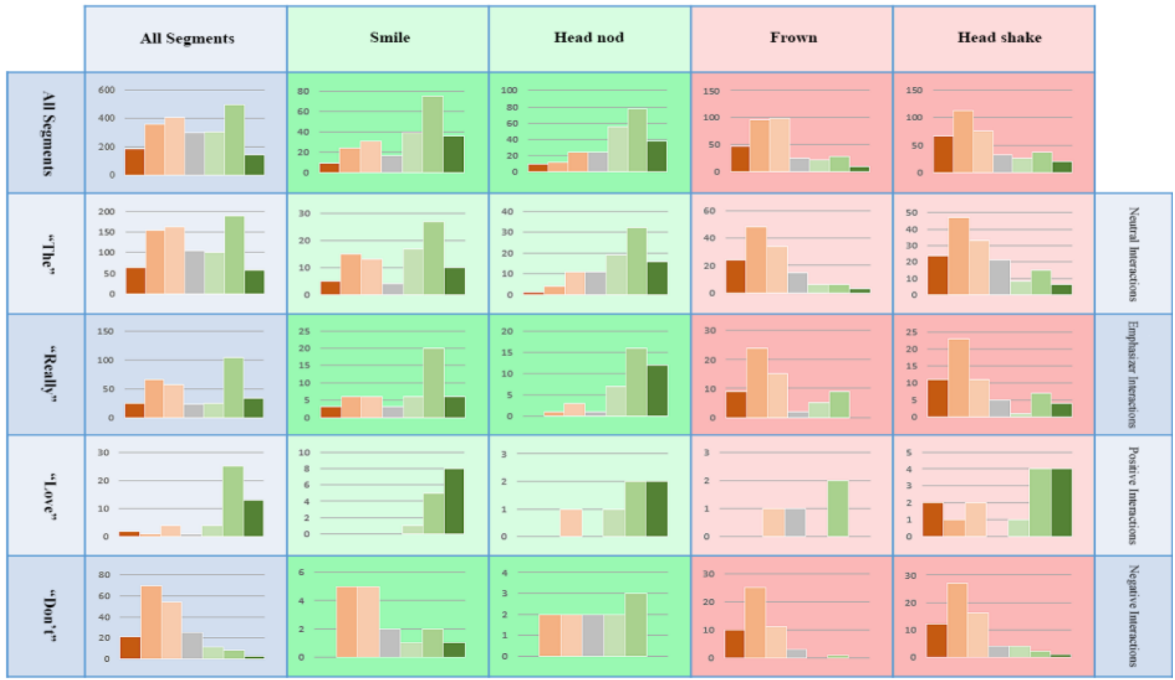

**Figure 3**: Sentiment intensity histograms for different spoken words and visual gestures. In each histogram y-axis is the frequency of co-occurrence and x-axis is sentiment intensity as in Figure 2.

| Total number of segments | 3702 |
| --- | --- |
| Total number of opinion segments | 2199 |
| Total number of objective segments | 1503 |
| Total number of videos | 93 |
| Total number of distinct speakers | 89 |
| Average number of opinion segments in video | 23.2 |
| Average length of opinion segments | 4.2 sec. |
| Average word count per opinion segments | 12 |
| Total number of words in opinion segments | 26,295 |
| Total of unique words in opinion segments | 3,107 |
| Total number of words in opinion segments appearing at least 10 times in the dataset | 557 |

**Table 1**: MOSI dataset statistics

Furthermore, manual gesture annotations are provided to study the relations between words and gestures. Since hands were not always visible in the YouTube videos, we decided to focus on facial gestures. We selected four gestures and expressions: smile, frown, headnod and headshake. These are expressive of emotions and regularly happen in MOSI. The annotation was carried out by simply marking the opinion segment for having each one of these expressions. An expert coder manually annotated all 2199 video segments and a second coder annotated a subset of this dataset to confirm a high coder agreement. For all 4 gestures the average coder agreement was 80.8%.

### 3.5 Multimodal analysis of words and visual gestures

MOSI dataset also enables detailed statistical study of language as a multimodal signal. The alignment between text, audio and video enables such analysis even at phoneme level. In this section we present a study to find a suitable multimodal representation for sentiment analysis. We want to understand the interaction patterns between spoken words and visual gestures. To study these interaction patterns, we propose to study the changes in the distribution of perceived sentiment intensity when a specific facial gesture is present or not. Our main research question can be rephrased as whether all spoken words are interacting similarly with facial gestures or if there are prototypical patterns in these multimodal interactions? This analysis was performed at the opinion-segment level where we studied the multimodal interactions of the top 100 spoken words with all 4 facial gestures (smile, head nod, frowning and head shake).

Figure 3 shows representative examples from our multimodal analysis where we identified four different types of interaction patterns between spoken words and facial gestures: neutral, emphasizer, positive and negative patterns. Each subgraph shown in Figure 3 is a histograms representing the distribution of perceived sentiment intensities per opinion segments.

To help understand the average interaction of facial gestures on spoken words, the first row of Figure 3 shows how the sentiment intensities are distributed for all opinion segments (the top left histogram of Figure 3 is repeated from Figure 2). These histograms are regarded as the common effect of gestures on spoken words. It is not surprising to see that opinion segments with a smile or a head nod are perceived as more positive. The opposite effect is observed for frown and head shake gestures.

**Neutral Interaction Pattern** To exemplify the neutral interaction pattern, we selected the most frequent word in our dataset, the word "the". "The" is considered a sentimentally neutral word since it is not positive or negative. The second row of Figure 3 shows the interaction between the facial gestures and the spoken word "the". We can observe that the pattern is mostly following the common interaction patterns in the first row (for all opinion segments).

**Emphasizer Interaction Pattern** A second interaction pattern was observed in our multimodal analysis. To exemplify this pattern, we showed in the third row of Figure 3 how facial gestures are interacting with the word "really". When accompanied by a smile or head nod, the distribution tends to shift to positive sentiment intensity, with less negative or neutral intensities. The opposite effect happens when it is accompanied by a frown or head shake, where the distribution is biased toward negative sentiment. In other words this interaction patterns tends to shift the sentiment towards the extremes. We define this type of interaction pattern as emphasizer.

**Negative or Positive Interaction Patterns** A third and fourth type of interaction seem to appear when studying sentimentally polarized words since their sentiment distribution are not affected the same way as the neutral and emphasizer. For example, the positively polarized word "love" is shown in the fourth row of Figure 3 (we merged the words "love" and "loved" in these histograms for simplification). We observe that the sentiment distributions do not significantly change polarity when accompanied by frown or head shake. An opposite trend happens when we study a negatively polarized word such as "don't" shown in the last row of Figure 3 (all the instances of "don't", "doesn't" and "didn't" have been merged in these histograms). We observe limited changes in the sentiment distributions for the smile and head nod. These are examples of positive and negative interaction patterns between words and facial gestures.

Based on these interaction patterns between words and gestures, we present a simple representation model that jointly takes into account words and gestures in each opinion segment. The detailed procedure to build this representation is presented in the next section.

## 4 Sentiment and Subjectivity Analysis Baselines

We designed a set of experiments to present baselines for sentiment and subjectivity analysis on MOSI dataset. All the experiment are done in a speaker independent framework; opinion segments of each speaker is either in training, validation or testing set (i.e. each speaker is only present in one of the sets; this is because models trained and tested on the same set of speakers are not generalizable to unseen speakers).

### 4.1 Sentiment Analysis Baselines

**Methodology:** All prediction models were trained using nu-SVR (Smola and Schölkopf, 2004; Chang and Lin, 2011) and tested using 5-fold cross-validation methodology. The automatic validation of the hyper-parameters was performed with 4-fold cross-validation on the training sets. The hyper parameters of linear nu-SVR are C and nu. In validation phase, C was automatically selected from powers of 10 in range [-5,3] and nu from [0.1,1] in steps of 0.1. The performance of the regressors is calculated based on mean absolute error (MAE) and correlation. In these studies we train the following models:

*Random*[7] We included in our experiments a simple baseline model which always predicts a random sentiment intensity between [3,-3]. This baseline gives an overall idea about how random models will work.

*Verbal* The second model was trained using only verbal features from MOSI. A very simple bag of words feature set was created from monograms and bigrams created from words in speech segments, including speech pause and pause fillers. All the features with less than 10 instances in the dataset were removed from the bag of words set given their infrequency.

---

[7] A baseline model that is slightly smarter than random would always predict the average intensity of training samples. This will give MAE of 1.39.

| | Spoken words | Verbal-only prediction | Visual gestures | Visual-only prediction | Multimodal Dictionary prediction | Ground Truth annotation |
|---|---|---|---|---|---|---|
| 1 | And quite honestly I wish I've seen this over the summer. | 0.14 | Smile, Head nod | 1.97 | 1.4 | 1.6 |
| 2 | Now I'm not gonna lie there're a few parts that have great action sequences now even though it is an animated film it did have some great fight scenes. | 2.3 | Head shake | -0.77 | 1.44 | 1.4 |

Table 3: Examples from dataset with predicted intensities for different models: verbal, visual and multimodal dictionary.

*Visual* The third model was trained using facial gestures described in Section 4.1. A binary feature is assigned for each of the 4 facial gesture: smile, frown, head nod and head shake.

*Verbal + Visual* The fourth model was trained on verbal and visual data combined. The verbal and visual features were simply concatenated for each opinion segment.

*Multimodal Dictionary* The fifth model is trained on joint representation of words and gestures. The procedure to build the multimodal dictionary is as follows: for each word $W_i$ from verbal features and gesture $G_j$, the set $\{(W_i \& G_j), (W_i \& \sim G_j)\}$ is added to the multimodal dictionary. ($W_i \& G_j$) captures the co-occurrence of word $W_i$ and gesture $G_j$. ($W_i \& \sim G_j$) captures the occurrences of word $W_i$ that did not occur with $G_j$. If both of them are present in the speech segment, the value of ($W_i \& G_j$) will be equal to 1. If only word $W_i$ is present without gesture $G_j$, the value of ($W_i \& \sim G_j$) will be equal to 1.

*Human Baseline* Humans are asked to predict the sentiment score of each opinion segment. This will be a baseline for how well humans can predict sentiment intensity and also a target for machine learning methods to reach in future.

| | MAE | Correlation |
|---|---|---|
| Random | 1.88 | 0 |
| Visual | 1.24 | 0.36 |
| Verbal | 1.18 | 0.46 |
| Verbal+Visual | 1.14 | 0.49 |
| Multimodal Dictionary | **1.1** | **0.53** |
| Human Performance | 0.61 | 0.83 |

Table 2: Mean absolute error and correlation for each of the trained baseline models.

| | SVM | DNN |
|---|---|---|
| Random | 0.59 | |
| Acoustic | 0.57 | 0.51 |
| Visual | 0.61 | 0.53 |
| Verbal | 0.65 | 0.57 |
| Multimodal Dictionary | **0.71** | **0.66** |

Table 4: subjectivity analysis accuracy of SVM and DNN.

The results sentiment analysis baselines are in table 2. As it can be seen multimodal dictionary built to model words and gestures jointly is outperforming simple concatenation of features (early fusion used in most of the previous works in multimodal sentiment analysis).

### 4.2 Subjectivity Analysis Baselines

The results for subjectivity analysis following two methods are presented here 1) linear C-SVM and 2) deep neural network. The training procedure for subjectivity uses information from automatically extracted audio visual features as well. Linear SVM hyperparameters are validated the same way as sentiment studies (section 4.1). Deep neural networks model is fully connected network with number of layers validated between [1, 5] and number of neurons in each layer validated from [10,50] in steps of 10. Table 4 presents the results of different baseline models for multimodal subjectivity analysis.

## 5 Discussion

Our experimental results in Table 2 show how combining verbal and visual cues help better predict sentiment. The multimodal dictionary built based on analysis from Section 3.5 has the best performance among all baselines. Table 3 shows examples on how information from visual gestures is helping the multimodal dictionary make more accurate predictions for sentiment. In the first case, it can be seen that verbal prediction is neutral while strong positive visual cues help multimodal dictionary to predict the intensity of the opinion more accurately. In the second case while verbal cues indicate highly positive, showing negative emotions carried by a headshake is an indicator that the opinion should not be considered highly positive. The results for sentiment and subjectivity analyses shown in Tables 2 and 4 highlight how using multimodal information helps in these tasks but the human performance results show that there is room for future research in these directions.

## 6 Conclusion

This paper introduces the MOSI dataset for multimodal sentiment intensity and subjectivity analysis. The dataset is the first multimodal sentiment analysis dataset with sentiment intensity and subjectivity annotations at opinion level. It has manual and automatic annotations of text, visual and audio features. Alignment between modalities opens the door to future multimodal studies of language. A new representation that captures the join distribution of words and gestures is presented based on statistical observations on the dataset. It is shown that using information more than just text can help models make better sentiment intensity predictions. The same is also true for subjectivity. We hope that this dataset opens the door to more detailed studies of sentiment and subjectivity analysis in multimedia content. Finally, the dataset is publicly available for download with all the extracted features.